\documentclass[letterpaper]{article} 
\usepackage{aaai23}  
\usepackage{times}  
\usepackage{helvet}  
\usepackage{courier}  
\usepackage[hyphens]{url}  
\usepackage{graphicx} 
\usepackage{natbib}  
\usepackage{caption} 
\usepackage{algorithm}
\usepackage{algorithmic}
\usepackage{amsthm}
\usepackage{amssymb}
\usepackage{amsfonts}
\usepackage{amsmath}

\usepackage{multirow}
\usepackage{xcolor}

\newcommand{\eat}[1]{}
\newcommand{\etal}{{et al.,~}}       
\newcommand{\ie}{{i.e.,~}}           
\newcommand{\wrt}{{w.r.t.,~}}         
\newcommand{\aka}{{a.k.a.,~}}        

\def\astrightarrow{\put(0.2,-2.2){*}\rightarrow}
\def\astleftarrow{\leftarrow\put(-5,-2.2){*}}

\newtheorem{definition}{Definition}

\newtheorem{theorem}{Theorem}

\usepackage{subcaption}
\makeatletter
\newcommand*{\indep}{%
	\mathbin{%
		\mathpalette{\@indep}{}%
	}%
}
\newcommand*{\nindep}{%
	\mathbin{
		\mathpalette{\@indep}{\not}
	}%
}
\newcommand*{\@indep}[2]{%
	\sbox0{$#1\perp\m@th$}
	\sbox2{$#1=$}
	\sbox4{$#1\vcenter{}$}
	\rlap{\copy0}
	\dimen@=\dimexpr\ht2-\ht4-.2pt\relax
	\kern\dimen@
	{#2}%
	\kern\dimen@
	\copy0 
}
\usepackage{newfloat}
\usepackage{listings}
\DeclareCaptionStyle{ruled}{labelfont=normalfont,labelsep=colon,strut=off} 
\floatstyle{ruled}
\newfloat{listing}{tb}{lst}{}
\floatname{listing}{Listing}
\title{Causal Inference with Conditional Instruments using Deep Generative Models}
\author{
    Debo~Cheng\textsuperscript{\rm 1,2}\equalcontrib, Ziqi~Xu\textsuperscript{\rm 2}\equalcontrib, Jiuyong~Li\textsuperscript{\rm 2}, Lin~Liu\textsuperscript{\rm 2}, Jixue~Liu\textsuperscript{\rm 2} and~Thuc~Duy~Le\textsuperscript{\rm 2} }
\affiliations{
    \textsuperscript{\rm 1} School of Computer Science and Engineering, Guangxi Normal University, Guilin, 541004, China\\

   \textsuperscript{\rm 2} UniSA STEM, University of South Australia, Adelaide, SA, 5095, Australia

    \{debo.cheng,ziqi.xu\}@mymail.unisa.edu.au, \{jiuyong.li,lin.liu,jixue.liu,thuc.le\}@unisa.edu.au
}

\usepackage{bibentry}

\begin{document}

\maketitle

\begin{abstract}
The instrumental variable (IV) approach is a widely used way to estimate the causal effects of a treatment on an outcome of interest from observational data with latent confounders. A standard IV is expected to be related to the treatment variable and independent of all other variables in the system. However, it is challenging to search for a standard IV from data directly due to the strict conditions. The conditional IV (CIV) method has been proposed to allow a variable to be an instrument conditioning on a set of variables, allowing a wider choice of possible IVs and enabling broader practical applications of the IV approach. Nevertheless, there is not a data-driven method to discover a CIV and its conditioning set directly from data. To fill this gap, in this paper, we propose to learn the representations of the information of a CIV and its conditioning set from data with latent confounders for average causal effect estimation. By taking advantage of deep generative models, we develop a novel data-driven approach for simultaneously learning the representation of a CIV from measured variables and generating the representation of its conditioning set given measured variables. Extensive experiments on synthetic and real-world datasets show that our method outperforms the existing IV methods.
\end{abstract}
	
\section{Introduction}
	\label{Sec:Intro}
	Estimating the causal effect of a treatment (\aka intervention, exposure, or action) on an outcome of interest, is a fundamental area of research~\citep{pearl2009causality,pearl2018book}. Randomised controlled trials (RCTs) are considered the gold standard for causal effect estimation. However, RCTs are often difficult or impossible to conduct due to ethical issues and/or high costs~\citep{imbens2015causal}. Thus, it is important to estimate causal effects from observational data. For the case when there are no latent or unmeasured confounders\footnote{A confounder is a variable that causally affects both the treatment and the outcome.} (\ie the unconfoundedness assumption~\cite{imbens2015causal} holds), many methods have been developed for causal effects estimation from data~\cite{yao2021survey,guo2020survey}. In contrast, for the more realistic and challenging case when there are latent confounders in data, only a handful of data-driven methods have been developed.

	Two types of approaches, the instrumental variable (IV) approach~\cite{hernan2006instruments} and the proxy approach~\cite{kuroki2014measurement,miao2018identifying} can be used for estimating  causal effects from data with latent confounders. The proxy approach relies on the assumption that the measured variables are noisy measurements of the latent confounders. However, recent work~\cite{rissanen2021critical} has shown that it is difficult to recover a blocking set (\ie a set that satisfies the back-door criterion~\cite{pearl2009causality}) for removing the confounding bias of the latent confounders even though there exists a rich set of proxy variables. Instead, the IV approach aims to avoid the spurious associations caused by latent confounders with the aid of a valid IV. In this work, we explore the direction of the IV approach.
	
	The traditional IV approach requires a standard IV (denoted as $S$ here) which satisfies the following three conditions~\cite{hernan2006instruments}: (i) $S$ is correlated with the treatment $T$ (\aka \textit{relevance condition}), (ii) $S$ affects the outcome $Y$ only through the treatment $T$ (\aka \textit{exclusion restriction}), and (iii) there is no confounding bias between $S$ and $Y$ (\aka \textit{unconfounded instrument}). However, the last two conditions are too strict to be satisfied in practice. A conditional IV (CIV, see Definition~\ref{def:conditionalIV} for its formal definition), which requires some measured variables as its conditioning set, has more relaxed conditions than a standard IV~\cite{brito2002generalized}. Hence in this paper, we focus on data-driven methods based on CIVs. 
	
\begin{figure}[t]
\centering
\includegraphics[scale=0.38]{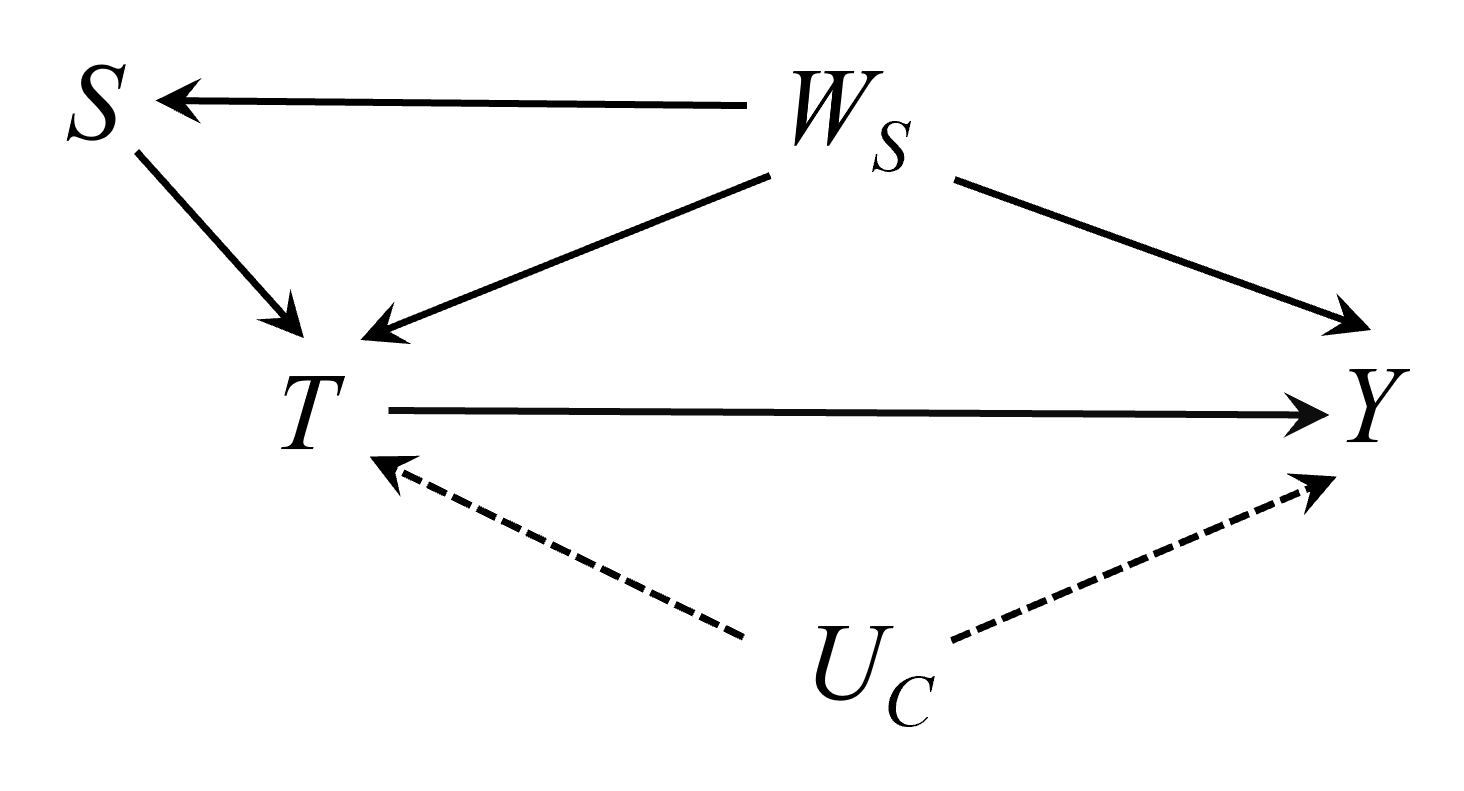}
\caption{An exemplar causal DAG shows the indistinguish ability between a CIV and its conditioning set in the data with a latent confounder $U_C$. $T$ and $Y$ are treatment and outcome variables. $S$ and $W_S$ are a CIV and its conditioning set, and they are indistinguishable in the data since both are dependent on $T$ and $Y$ and conditional dependent on $Y$ given $T$. Dashed edges indicate that $U_C$ cannot be measured. As from the DAG, we know that in the dataset generated, both $S$ and ${W}_S$ are associated with $T$, given any other observed variable(s). Moreover, while $S$ is associated with $Y$ given any other observed variable(s), ${W}_S$ is also associated with $Y$ given any other observed variable(s) due to the unmeasured confounder ${U}_{C}$.} 
\label{fig:observedvariables}
\end{figure}

	It is challenging to determine a CIV and its conditioning set directly from observational data with latent confounders since a CIV and its conditioning set are not distinguishable by statistical tests due to latent confounders~\cite{spirtes2000causation,pearl2009causality}. The example in Fig.~\ref{fig:observedvariables} illustrates the challenge. The causal directed acyclic graph (DAG) in Fig.~\ref{fig:observedvariables} shows an underlying data generation mechanism, where $T$ and $Y$ are the treatment and outcome variables respectively, ${U}_{C}$ is a latent confounder, and $S$ is a valid CIV for which ${W}_S$ is the conditioning set. In a dataset generated from this mechanism, it is impossible to distinguish the roles of $S$ (as a CIV) and ${W}_S$ (as the conditioning set of $S$). 
	
	Recently, the deep generative model based on variational autoencoder (VAE)~\cite{kingma2014auto,sohn2015learning} has achieved many successes in areas such as causal representation learning~\cite{scholkopf2021toward,scholkopf2022causality} and individual causal effect estimation~\cite{hassanpour2019learning,zhang2021treatment}. In this work, we will leverage the strength of VAE to address the challenge of discovering CIVs and their conditioning sets from data with latent confounders.
 
	For a treatment $T$ and outcome $Y$, and a set of measured pretreatment variables $\mathbf{X}$, and given the assumption that there exists at least one CIV in $\mathbf{X}$, we propose the causal representation learning scheme (as shown in Fig.~\ref{fig:desiredDAG}) to learn the representation $\mathbf{Z}_T$ of $\mathbf{X}$ and generate the latent representation $\mathbf{Z}_\mathbf{C}$ conditioning on $\mathbf{X}$ respectively, and we prove that the obtained $\mathbf{Z}_T$ is a valid CIV conditioning on $\mathbf{Z}_\mathbf{C}$. We then develop a VAE-based method, named CIV.VAE (Conditional IV approach based on VAE model) to conduct causal representation learning to obtain $\mathbf{Z}_T$ and $\mathbf{Z}_\mathbf{C}$ for unbiased average causal effect estimation.  
	
The contributions of this work are summarised as follows.
\begin{itemize}
		\item We propose to tackle the problem of discovering a CIV and its conditioning set from data in the presence of latent confounders with causal representation learning.  As far as we know, this is the first work using a data-driven approach and  causal representation learning for identifying CIVs and their conditioning sets. 
		\item We develop a novel VAE based method, CIV.VAE, to learn the representation of CIVs and generate the representation of their conditioning sets for causal effect estimation from data in the presence of latent confounders.
		\item Extensive experiments on a wide range of synthetic and real-world datasets show that the causal effects estimated using the CIVs and conditioning sets obtained by CIV.VAE have the smallest estimation error compared with the state-of-the-art causal effect estimators.
	\end{itemize}

The rest of the paper is organised as follows. We firstly introduce background knowledge in Preliminary. Secondly, the details of our proposed CIV.VAE method is presented. Thirdly, we discuss the experimental setup, datasets and experimental results. Fourthly, we review the related works. Finally, we conclude our work. 

\section{Preliminary}
\label{sec:pre}
In this section, we briefly introduce some important notations, definitions and assumptions used in the paper.

\subsection{Notations \& Definitions}
\label{subsec:notations}
We use uppercase and lowercase letters to represent variables and their values, respectively. Bold-faced uppercase and lowercase letters are used to indicate a set of variables and a value assignment of the set, respectively. 

Let $T$ be a binary treatment variable ($T=1$ for treated and $T=0$ for control) and $Y$ be the outcome of interest. Let $\mathcal{G}=(\mathbf{V, E})$ be a DAG with nodes $\mathbf{V}= \mathbf{X}\cup\mathbf{U}\cup\{T, Y\}$ and directed edges $\mathbf{E}$. $\mathbf{X}$ is the set of measured pretreatment variables, and  $\mathbf{U}=\mathbf{U}_\mathbf{C}\cup\mathbf{U}'$ is the set of unmeasured confounders, where $\mathbf{U}_\mathbf{C}$ denotes the latent confounders between the pair $T$ and $Y$, and $\mathbf{U}'$ represents the latent confounders between other variable pairs. In a causal DAG  $\mathcal{G}=(\mathbf{V, E})$, an edge (directed) represents the causal relationship between two nodes. For example, $X_i\rightarrow X_j$ in $\mathcal{G}$ indicates that $X_i$ is a cause of $X_j$, and $X_j$ is an effect of $X_i$. In a DAG $\mathcal{G}$, a path between $X_{i}$ and $X_{j}$ consists of a sequence of distinct nodes $\langle X_{i}, \dots, X_{j}\rangle$ with every pair of successive nodes being adjacent. More definitions regarding graphical causal modelling such as $d$-separation, collider, Markov property and faithfulness can be found in the supplement.

In this work, we would like to query the average causal effect ($ACE$) of treatment variable $T$ on outcome variable $Y$, referred to as $ACE(T, Y)$, from a dataset containing a set of pretreatment variables $\mathbf{X}$, treatment $T$ and outcome $Y$, and assuming that $\mathbf{X}$ contains at least one CIV and its conditioning set and there exists a set of latent confounders $\mathbf{U}_\mathbf{C}$ affecting both $T$ and $Y$ and $\mathbf{U}_\mathbf{C}\neq \emptyset$. There may exist other latent confounders (denoted as $\mathbf{U}'$) which affect pairs of variables other than $(T, Y)$, and $\mathbf{U}'$ can be an $\emptyset$.

\subsection{Conditional Instrumental Variable (CIV)}
\label{subsection:CIV}
The IV approach is a powerful method for removing the confounding bias caused by the latent confounders affecting both treatment and outcome. As discussed in the Introduction, the last two conditions of a standard IV are too strict to be satisfied in real-world applications. In contrast, a conditional IV (CIV) , as defined below, has more relaxed conditions than a standard IV~\cite{brito2002generalized,pearl2009causality}. 

\begin{definition}[Conditional IV~\cite{pearl2009causality}]
	\label{def:conditionalIV}
	Given a DAG $\mathcal{G}=(\mathbf{V, E})$ with $\mathbf{V}=\mathbf{X}\cup\mathbf{U}\cup\{T, Y\}$, a variable $S$ is said to be a CIV \wrt $T\rightarrow Y$ if there exists a set of measured variables $\mathbf{W}\subseteq\mathbf{X}\setminus\{S\}$ and $\mathbf{W}\neq \emptyset$ such that (i) $S\nindep_d T|\mathbf{W}$, (ii) $S\indep_d Y|\mathbf{W}$ in $\mathcal{G}_{\underline{T}}$,  where $\mathcal{G}_{\underline{T}}$ is a manipulated DAG obtained by removing $T\rightarrow Y$ from $\mathcal{G}$, and (iii) $\forall W\in \mathbf{W}$, $W$ is not a descendant of $Y$.
\end{definition}
In the above definition, $\nindep_d$ and $\indep_d$ indicate d-connection and d-separation respectively in a DAG~\cite{pearl2009causality}. Definition~\ref{def:conditionalIV} can be easily generalised to a set of CIVs $\mathbf{S}$ with the corresponding conditioning set $\mathbf{W}$. Under the pretreatment assumption (i.e. all variables in $\mathbf{X}$ are pretreatment variables), condition (iii) is always satisfied, so we only need to test conditions (i) and (ii). Given $X\in \mathbf{X}$, if a set $\mathbf{W}\subseteq\mathbf{X}\setminus\{X\}$ is found to block all paths between $X$ and $Y$; and given $\mathbf{W}$, $X$ and $T$ are still dependent, then $\mathbf{W}$ instrumentalises $X$ to be a CIV. Note that finding such a $\mathbf{W}$ from a causal DAG is proved to be NP-complete even with the  pretreatment assumption~\cite{van2015efficiently}.

 In this work, we utilise the two stage least squares method~\cite{angrist1995two} as the ACE estimator. In a linear system, once we have a valid CIV $S$ and its conditioning set $\mathbf{W}$, $ACE(T, Y)$ can be estimated by $\sigma_{s*y*\mathbf{w}}/\sigma_{s*t*\mathbf{w}}$, where $\sigma_{s*y*\mathbf{w}}$ and $\sigma_{s*t*\mathbf{W}}$ are the estimated causal effect of $S$ on $Y$ conditioning on $\mathbf{W}$ and the causal effect of $S$ on $T$ conditioning on $\mathbf{W}$, respectively. 

\section{The Proposed CIV.VAE Model}
 \label{sec:civvae}
\subsection{The Causal Representation Learning Scheme for CIV Discovery}
 \label{subsec:theorem}
\begin{figure}[t]
	\centering
	\includegraphics[scale=0.38]{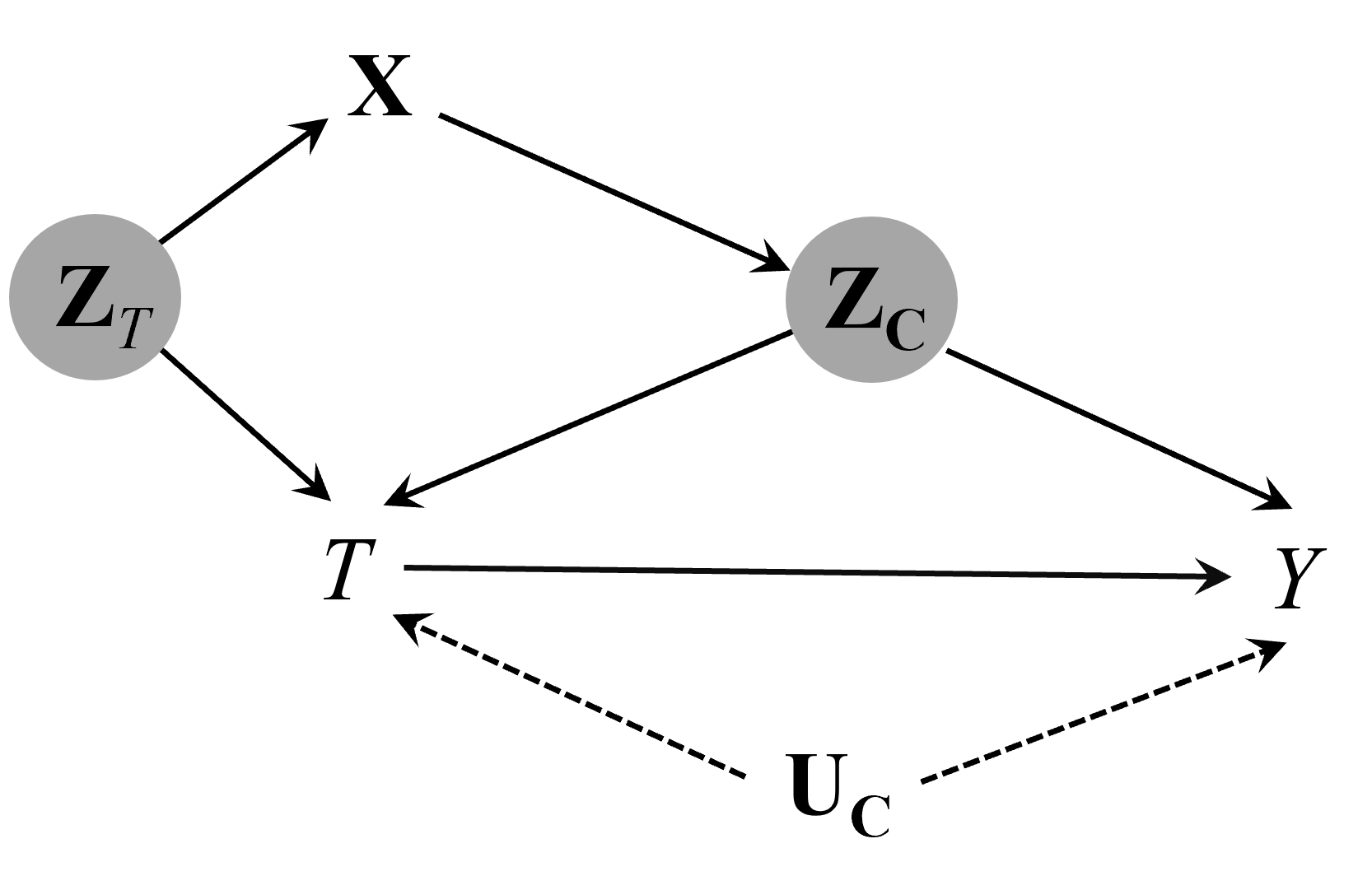}
	\caption{A causal graph representing the proposed causal representation learning scheme for discovering CIVs and their conditioning sets. $T$, $Y$, $\mathbf{X}$, and $\mathbf{U}_\mathbf{C}$ are the treatment, the outcome, the set of measured pretreatment variables, and latent confounders between $T$ and $Y$, respectively. The two grey circles denote the representation $\mathbf{Z}_{T}$ containing the CIV information in $\mathbf{X}$ and $\mathbf{Z}_{\mathbf{C}}$ holding the information of the conditioning set of $\mathbf{Z_T}$ generated given $\mathbf{X}$. } 
	\label{fig:desiredDAG}
\end{figure}

In this work, we aim to simultaneously learn the latent representation $\mathbf{Z}_T$ of $\mathbf{X}$ and generate the latent representation $\mathbf{Z}_\mathbf{C}$ given $\mathbf{X}$. Here $\mathbf{Z}_{T}$ contains the instrumental information that only influences $T$ but not $Y$, and $\mathbf{Z}_{\mathbf{C}}$ denotes the representation of the confounders that affect both $T$ and $Y$. We assume that Fig.~\ref{fig:desiredDAG} is the underlying generative model (\ie the underlying causal DAG) and it shows our proposed causal representation learning scheme. Specifically, a set of pretreatment variables $\mathbf{X}$ are generated from $\mathbf{Z}_{T}$ and $\mathbf{Z}_{T}$ captures the information of $\mathbf{S}$ in Fig.~\ref{fig:observedvariables}. The latent confounding representation $\mathbf{Z}_{\mathbf{C}}$ is generated based on $\mathbf{X}$ and captures the information of $\mathbf{X}\setminus\mathbf{S}$ in Fig.~\ref{fig:observedvariables}. 

 If the two disjoint representations $\mathbf{Z}_{T}$ and $\mathbf{Z}_{\mathbf{C}}$ can be inferred from data with latent confounders, the following proposed theorem guarantees that $\mathbf{Z}_{T}$ is a valid CIV with $\mathbf{Z}_{\mathbf{C}}$ being its conditioning set. Using $\mathbf{Z}_{T}$ and $\mathbf{Z}_{\mathbf{C}}$, we can obtain unbiased causal effect estimation from data with latent confounders.

\begin{theorem}
		\label{theorem:001}
	Given a causal DAG $\mathcal{G}\!\!=\!\!(\mathbf{X}\cup \mathbf{U}\cup \{T, Y\}, \mathbf{E})$ where $\mathbf{X}$ is the set of pretreatment variables. $T$ and $Y$ are the treatment and outcome respectively, and there exists $T\rightarrow Y$. $\mathbf{U}=\mathbf{U}_\mathbf{C}\cup\mathbf{U}'$ is the set of latent confounders between $T$ and $Y$ (denoted as $\mathbf{U}_\mathbf{C}$) and latent confounders between measured variables in $\mathbf{X}$ (denoted as $\mathbf{U}'$), and $\mathbf{E}$ is the set of edges between the variables. Suppose that there exists a set of CIVs $\mathbf{S}\subset\mathbf{X}$ with $\left |\mathbf{S}\right |\geq 1$ and its conditioning set $\mathbf{W}\subseteq\mathbf{X}\setminus\mathbf{S}$ with $\left |\mathbf{W}\right |\geq 1$. If the latent representations  $\mathbf{Z}_{T}$ and $\mathbf{Z}_{\mathbf{C}}$ as shown in Fig.~\ref{fig:desiredDAG} can be learned from data, then $\mathbf{Z}_{T}$ is a CIV conditioning on $\mathbf{Z}_{\mathbf{C}}$ for estimating the causal effect of $T$ on $Y$.
\end{theorem}
\begin{proof}
	We prove that the two representations $\mathbf{Z}_{T}$ and $\mathbf{Z}_{\mathbf{C}}$ satisfy the three conditions of Definition~\ref{def:conditionalIV} based on the causal DAG shown in Fig.~\ref{fig:desiredDAG}. The conditions $\left |\mathbf{S}\right |\geq 1$ and $\left |\mathbf{W}\right |\geq 1$ are to ensure that there exists at least a pair of CIV and its conditioning set in $\mathbf{X}$. In  $\mathcal{G}$, (1) $\mathbf{Z}_T$ is a set of parents of the treatment $T$, so $\mathbf{Z}_T \nindep_d T|\mathbf{Z}_{\mathbf{C}}$ and the first condition of Definition~\ref{def:conditionalIV} is satisfied; (2) the spurious association between $\mathbf{Z}_T$ and $Y$ are caused by these paths, $\mathbf{Z}_T \rightarrow \mathbf{X}\rightarrow \mathbf{Z}_\mathbf{C}\rightarrow Y$, $\mathbf{Z}_T \rightarrow T\leftarrow \mathbf{U}_\mathbf{C}\rightarrow Y$, $\mathbf{Z}_T \rightarrow T\leftarrow \mathbf{Z}_\mathbf{C}\rightarrow Y$ and $\mathbf{Z}_T \rightarrow \mathbf{X} \rightarrow \mathbf{Z}_\mathbf{C} \rightarrow T \leftarrow \mathbf{U}_\mathbf{C}\rightarrow Y$ in the manipulated DAG  $\mathcal{G}_{\underline{T}}$. The first path is blocked by $\mathbf{Z}_\mathbf{C}$ and the other three paths are blocked by $\emptyset$ since these paths contain a collider $T$, \ie  $\mathbf{Z}_T \indep_d Y| \mathbf{Z}_\mathbf{C}$ in  $\mathcal{G}_{\underline{T}}$ and the second condition of Definition~\ref{def:conditionalIV} holds; (3) $\mathbf{Z}_{\mathbf{C}}$ is generated based on $\mathbf{X}$ and $\mathbf{X}$ contains only pretreatment variables, so $\mathbf{Z}_{\mathbf{C}}$ is not a descendant of $Y$, \ie the third condition of Definition~\ref{def:conditionalIV} holds. Hence, the two representations $\mathbf{Z}_T$ and $\mathbf{Z}_{\mathbf{C}}$ satisfy Definition~\ref{def:conditionalIV}, \ie $\mathbf{Z}_{\mathbf{C}}$ instrumentalises $\mathbf{Z}_T$ such that $\mathbf{Z}_T$ is a valid CIV for estimating the causal effect of $T$ on $Y$. 
\end{proof}

Theorem~\ref{theorem:001} permits us to develop a data-driven method based on deep generative models~\cite{kingma2019introduction} to learn the representations of a CIV and its conditioning set directly from observational data. In the next subsection, we introduce our proposed data-driven method, CIV.VAE for learning the two representations $\mathbf{Z}_T$ and $\mathbf{Z}_{\mathbf{C}}$.

\subsection{VAE-based Representation Learning of $\mathbf{Z}_T$ and $\mathbf{Z}_{\mathbf{C}}$}
\label{sec:cvae}

\begin{figure*}[t]
		\centering
		\begin{subfigure}{.49\textwidth}
			\centering
			\includegraphics[scale=0.5]{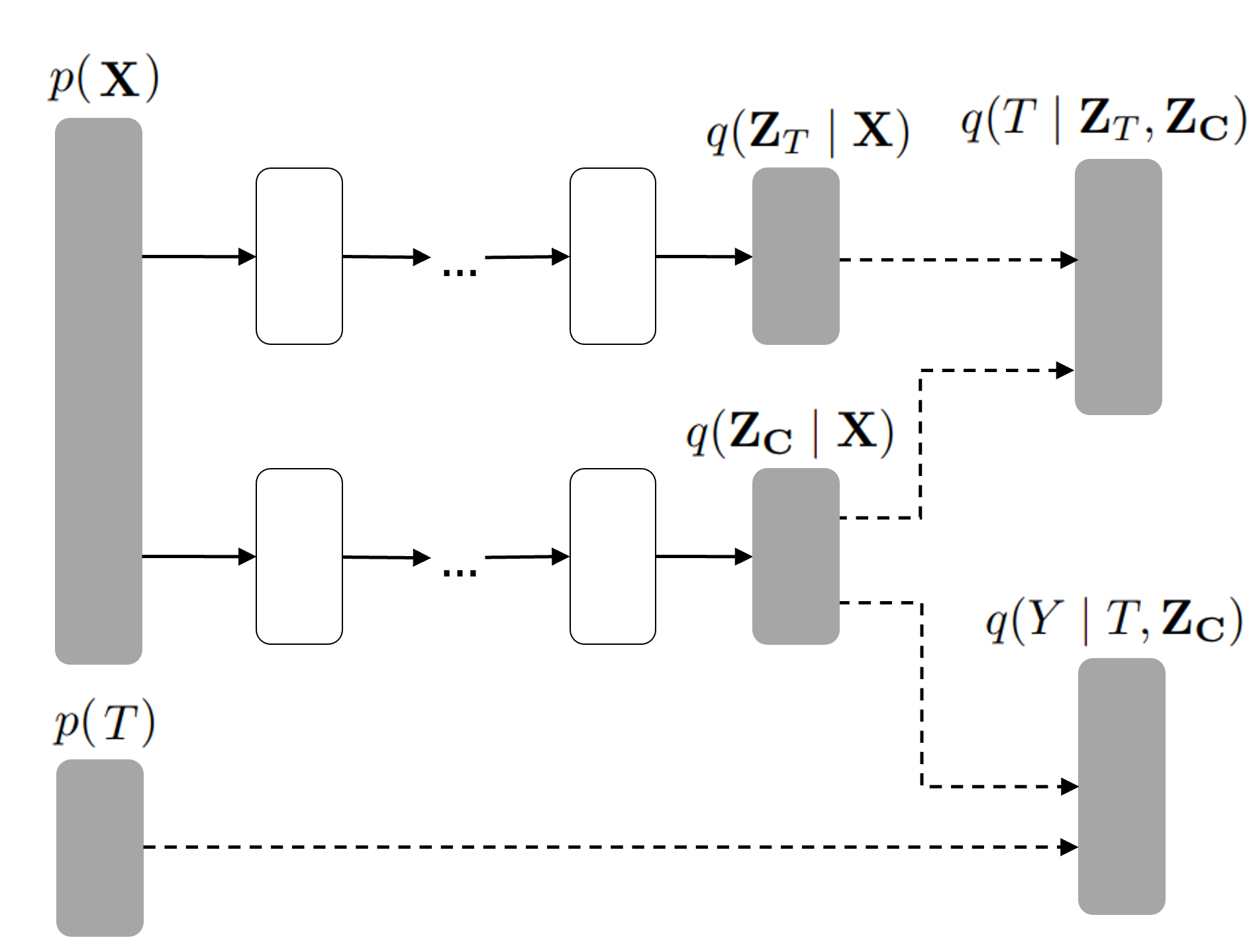}
			\caption{Inference Network}
			\label{fig:sub1}
		\end{subfigure}
		\begin{subfigure}{.49\textwidth}
			\centering
			\includegraphics[scale=0.5]{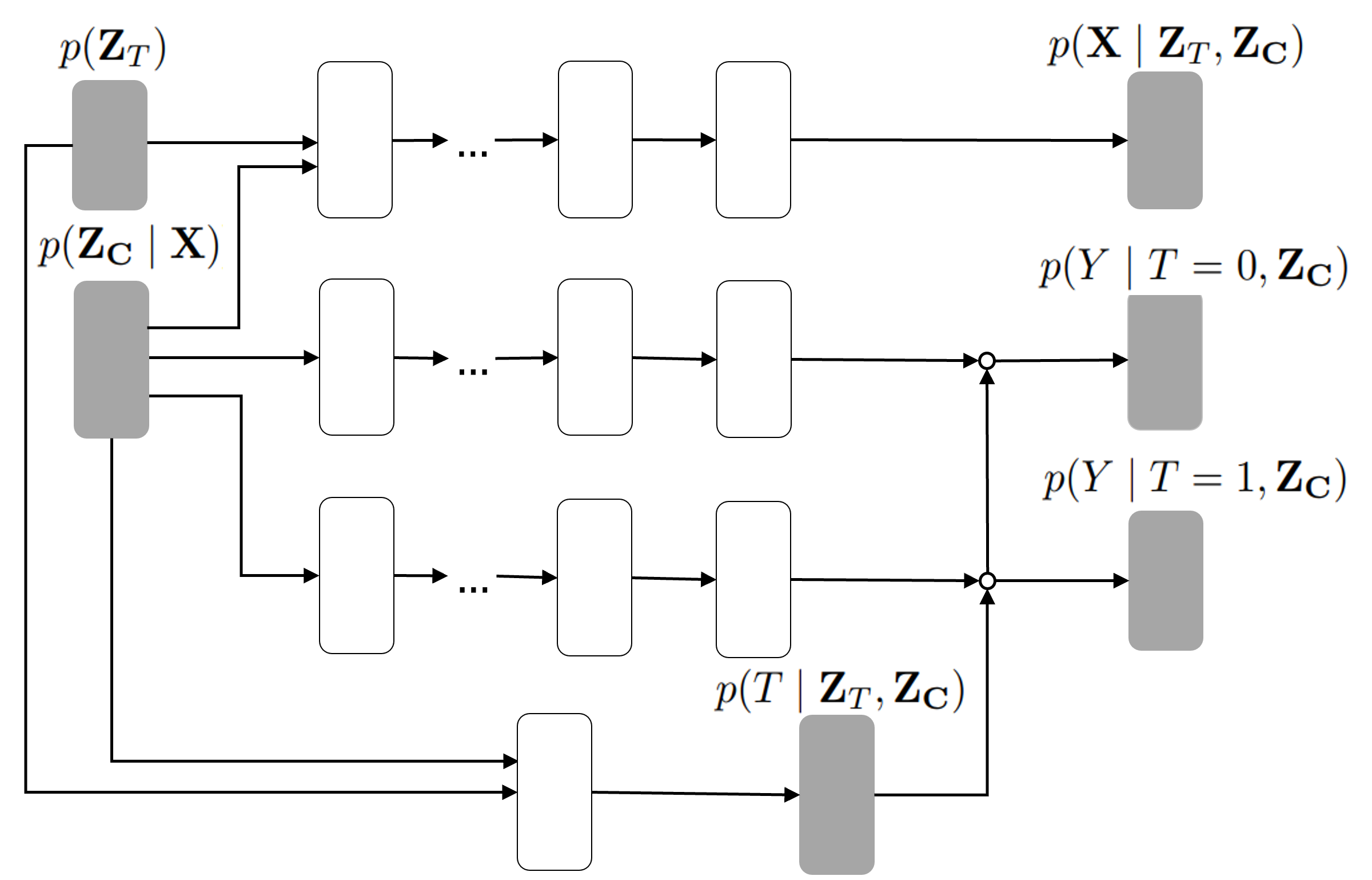}
			\caption{Generative Network}
			\label{fig:sub2}
		\end{subfigure}
		\caption{The proposed CIV.VAE architecture which consists of the inference network and the generative network for learning the latent representations of CIV $\mathbf{Z}_T$ and its conditioning set $\mathbf{Z}_{\mathbf{C}}$. A grey box denotes the drawing of samples from the respective distribution, a white box indicates the parameterised deterministic neural network transitions, and a circle indicates switching paths according to the value of $T$. In the inference network, the dashed arrows indicate the two auxiliary predictors $q(T|\mathbf{Z}_{T},\mathbf{Z}_\mathbf{C})$ and $q(Y| T, \mathbf{Z}_\mathbf{C})$.}
		\label{fig:architecture}
\end{figure*}

Fig.~\ref{fig:architecture} shows the CIV.VAE architecture we have developed for learning the latent representations of the CIV $\mathbf{Z}_T$ and its conditioning set $\mathbf{Z}_\mathbf{C}$. CIV.VAE comprises an inference network and a generative network as shown in Fig.~\ref{fig:architecture}(a) and Fig.~\ref{fig:architecture}(b) respectively. CIV.VAE utilises the inference network and the generative network to approximate the posterior distributions of $p(\mathbf{Z}_T|\mathbf{X})$ and $p(\mathbf{Z}_\mathbf{C}|\mathbf{X})$ for the two latent representations $\mathbf{Z}_T$ and $\mathbf{Z}_{\mathbf{C}}$ which indicate the latent CIV representation and the representation of its conditioning set, respectively.

In the inference network, two separate encoders $q(\mathbf{Z}_{T}|\mathbf{X})$ and $q(\mathbf{Z}_\mathbf{C}|\mathbf{X})$ are utilised as variational posteriors over the latent representations. In the generative model, these latent representations are utilised by a single decoder $p(\mathbf{X}|\mathbf{Z}_{T},\mathbf{Z}_\mathbf{C})$ for the reconstruction of $\mathbf{X}$. Based on the standard VAE framework in literature~\cite{kingma2014auto,kingma2019introduction}, the prior distribution of $p(\mathbf{Z}_{T})$ is sampled from a Gaussian distribution as follows:
\begin{equation}
	\label{eq:001}
	\begin{aligned}
		p(\mathbf{Z}_{T}) &= \prod_{i=1}^{D_{\mathbf{Z}_{T}}} \mathcal{N}(Z_{T_i}| 0, 1).
	\end{aligned}	
\end{equation}
\noindent where $D_{\mathbf{Z}_{T}}$ is the dimension of $\mathbf{Z}_{T}$.

Specifically, in the inference model, the variational approximations of the posterior distributions are as follows:
\begin{equation}
	\label{eq:002}
	\begin{aligned}
		q(\mathbf{Z}_{T}|\mathbf{X}) &= \prod_{i=1}^{D_{\mathbf{Z}_{T}}} \mathcal{N}(\mu = \hat{\mu}_{\mathbf{Z}_{T_i}}, \sigma^2 = \hat{\sigma}^2_{\mathbf{Z}_{T_i}});\\
		q(\mathbf{Z}_\mathbf{C}|\mathbf{X}) &= \prod_{i=1}^{D_{\mathbf{Z}_\mathbf{C}}} \mathcal{N}(\mu = \hat{\mu}_{\mathbf{Z}_{\mathbf{C}_i}}, \sigma^2 = \hat{\sigma}^2_{\mathbf{Z}_{\mathbf{C}_i}})
	\end{aligned}	
\end{equation} 
\noindent where $\hat{\mu}_{\mathbf{Z}_{T}},\hat{\mu}_{\mathbf{Z}_\mathbf{C}}$ and $\hat{\sigma}^2_{\mathbf{Z}_{T}},\hat{\sigma}^2_{\mathbf{Z}_\mathbf{C}}$ are the means and variances of the Gaussian distributions parameterised by neural networks. $D_{\mathbf{Z}_\mathbf{C}}$ is the dimension of $\mathbf{Z}_\mathbf{C}$.

In the generative model, according to the conditional variational autoencoder (CVAE) network~\cite{sohn2015learning}, we use Monte Carlo (MC) sampling to obtain $\mathbf{Z}_\mathbf{C}$ conditioning on the set of pretreatment variables $\mathbf{X}$:
\begin{equation}
		\label{eq:003}
	\begin{aligned}
		\mathbf{Z}_\mathbf{C} \backsim p(\mathbf{Z}_\mathbf{C}|\mathbf{X})
	\end{aligned}	
\end{equation} 

The generative models for $T$ and $\mathbf{X}$ are described as:
\begin{equation}
		\label{eq:004}
	\begin{aligned}
		p(T|\mathbf{Z}_{T},\mathbf{Z}_\mathbf{C}) &= Bern(\sigma(g_1(\mathbf{Z}_{T}, \mathbf{Z}_\mathbf{C})));\\ p(\mathbf{X}|\mathbf{Z}_{T}, \mathbf{Z}_\mathbf{C}) &= \prod_{i=1}^{D_{\mathbf{X}}} p(X_i|\mathbf{Z}_{T}, \mathbf{Z}_\mathbf{C}),
	\end{aligned}	
\end{equation} where $g_1(\cdot)$ is the function parameterised by neural networks and $\sigma(\cdot)$ is the logistic function.

For continuous $Y$,  in the generative model, it is modelled as a Gaussian distribution with its mean and variance parameterised it by the mutually exclusive neural networks that defines $p(Y | T = 0, \mathbf{Z}_\mathbf{C})$ and $p(Y | T = 1, \mathbf{Z}_\mathbf{C})$, respectively. Specifically, the model is defined as:

\begin{equation}
		\label{eq:005}
	\begin{aligned}
		&p(Y | T,\mathbf{Z}_\mathbf{C}) = \mathcal{N}(\mu = \hat{\mu}_{Y}, \sigma^2 = \hat{\sigma}^2_{Y}),\\
		&\hat{\mu}_{Y} = T \cdot g_2(\mathbf{Z}_\mathbf{C}) + (1-T) \cdot g_3(\mathbf{Z}_\mathbf{C}),\\
		&\hat{\sigma}^2_{Y} = T \cdot g_4(\mathbf{Z}_\mathbf{C}) + (1-T) \cdot g_5(\mathbf{Z}_\mathbf{C}).
	\end{aligned}	
\end{equation}
\noindent where $g_2(\cdot), g_3(\cdot), g_4(\cdot)$ and $g_5(\cdot)$ are the functions parameterised by neural networks. For binary $Y$, we model it with a Bernoulli distribution. The specific model is:

\begin{equation}
	\label{eq:006}
	\begin{aligned}
		&p(Y| T, \mathbf{Z}_\mathbf{C}) = Bern(\sigma(g_6(T, \mathbf{Z}_\mathbf{C}))).
	\end{aligned}	
\end{equation}
\noindent where $g_6(\cdot)$ is a function parameterised by neural networks. For inference, the parameters can be optimised by maximising the evidence lower bound (ELBO):

\begin{equation}
	\label{eq:007}
	\begin{aligned}
		\mathcal{M} =~ &\mathbb{E}_{q}[\log p(\mathbf{X}| \mathbf{Z}_{T}, \mathbf{Z}_\mathbf{C})] - D_{KL}[q(\mathbf{Z}_{T}|\mathbf{X})||p(\mathbf{Z}_{T})]\\ &- D_{KL}[q(\mathbf{Z}_\mathbf{C}|\mathbf{X})||p(\mathbf{Z}_\mathbf{C}|\mathbf{X})].
	\end{aligned}	
\end{equation}

Note that the decoder (generative network) $p(\mathbf{Z}_\mathbf{C}|\mathbf{X})$ conditioning on $\mathbf{X}$ is used to encourage as much information as possible from $\mathbf{X}$ is captured in the CIV.VAE method. To improve the learning of the latent representations $\mathbf{Z}_{T}$ and $\mathbf{Z}_{\mathbf{C}})$ and enable that the treatment $T$ can be predicted from the two representations and the outcome $Y$ can be predicted from $\mathbf{Z}_\mathbf{C}$ and $T$, we add two auxiliary predictors to the variational ELBO in Eq.(\ref{eq:007}) as designed by the works~\cite{louizos2017causal,zhang2021treatment}. Consequently, the proposed objective function of CIV.VAE is expressed as:

\begin{equation}
	\label{eq:008}
	\begin{aligned}
		\mathcal{L}_{CIV.VAE} =~ &-\mathcal{M} + \alpha \mathbb{E}_{q}[\log q(T|\mathbf{Z}_{T}, \mathbf{Z}_\mathbf{C})]\\ &+ \beta \mathbb{E}_{q}[\log q(Y| T, \mathbf{Z}_\mathbf{C})],
	\end{aligned}	
\end{equation}
\noindent where $\alpha$ and $\beta$ are the weights for the auxiliary predictors.

	For calculating $ACE(W, Y)$, we draw $\mathbf{Z}_{T}$ and $\mathbf{Z}_\mathbf{C}$ from the trained CIV.VAE method, and utilise both learned latent representations $\mathbf{Z}_{T}$ and $\mathbf{Z}_\mathbf{C}$ in the Instrumental Variable (CIV) method~\cite{angrist1995two} as described in the ``Conditional Instrumental Variable (CIV)'' subsection, where $s$ and $w$ are replaced by $\mathbf{Z}_{T}$ and $\mathbf{Z}_\mathbf{C}$ respectively. 
	
	The main advantage of CIV.VAE is that it simultaneously learns the latent CIV representation $\mathbf{Z}_{T}$ and the latent representation of the conditioning set $\mathbf{Z}_{\mathbf{C}}$ for  $\mathbf{Z}_{T}$ without specifying a CIV and its conditioning set by domain knowledge, and it provides a practical solution to the challenge (described in the Introduction) of distinguishing a CIV and its conditioning set from data with latent confounders. Hence, CIV.VAE is expected to have wider applications. CIV.VAE only relies on two practical assumptions, the pretreatment variable assumption and the existence of at least one CIV and its conditioning set in $\mathbf{X}$.

 	The main difference between CIV.VAE and the two other VAE-based causal effect estimators, CEVAE~\cite{louizos2017causal} and TEDVAE~\cite{zhang2021treatment} is that CIV.VAE builds on conditioning VAE for learning $\mathbf{Z}_T$ as the CIV and $\mathbf{Z}_{\mathbf{C}}$ as its conditioning set that blocks the confounding bias between $\mathbf{Z}_{T}$ and $Y$, whereas CEVAE and TEDVAE are to recover the set that blocks the confounding bias between $T$ and $Y$. Moreover, CIV.VAE method belongs to the IV approach, while CEVAE and TEDVAE methods are confounding adjustment methods.  	

\section{Experiments}
\label{sec:exp}
In this section, we evaluate the performance of CIV.VAE for the task of estimating the average causal effect of $T$ on $Y$. The experiments are divided into two parts: evaluation with simulated data and evaluation with real-world data. For the first part, we use the causal DAG in Fig. 1 in the supplement to generate synthetic datasets with latent confounders. For the second part, we use three real-world datasets, Schoolingreturns~\cite{card1993using}, 401k~\cite{wooldridge2010econometric} and Sachs~\cite{sachs2005causal}, which have reference causal effect values available in literature. The three datasets are widely used in the evaluation of IV-based methods~\cite{abadie2003semiparametric,wooldridge2010econometric,silva2017learning}. Note that Schoolingreturns and 401k each have a nominated CIV for the causal effect estimation, but the corresponding conditioning set is unknown, and there is not a nominated IV in Sachs. Our CIV.VAE method do not use a known CIV, instead we learn the representations of CIVs and their conditioning sets.

\subsection{Experiment Setup}
We compare CIV.VAE with three types of causal effect estimators: (1) IV-based estimators with a given IV, (2) IV-based estimators without a given IV and (3) VAE-based causal effect estimators. Five of the IV-based estimators, TSLS (two-stage least squares) regression~\cite{angrist1995two}, FIVR (the causal random forest method for IV regression)~\cite{athey2019generalized}, the deepIV (a deep learning based IV estimator)~\cite{hartford2017deep}, OrthoIV (orthogonal machine learning based IV estimator)~\cite{syrgkanis2019machine} and DMLIV (double machine learning based IV estimator)~\cite{chernozhukov2018double}, each of which needs a given IV; whereas the other IV-based estimators, IV.Tetrad~\cite{silva2017learning} and sisVIVE (some invalid some valid IV estimator)~\cite{kang2016instrumental} do not need a given IV. The two VAE-based causal effect estimators are CEVAE (causal effect variational autoencoder)~\cite{louizos2017causal} and TEDVAE (treatment effect by disentangled variational autoencoder)~\cite{zhang2021treatment}. We choose the two VAE-based estimators as our baseline because CIV.VAE also builds on the VAE model. 

\paragraph{Evaluation Metrics.} For synthetic datasets with the true causal effect $ACE(T, Y)$, the absolute error: $\varepsilon_{ACE} = \left |\hat{ACE}(T, Y)-ACE(T, Y)\right|$ is used to evaluate the performance of all estimators. For multiple replications, we report the average results with STD (standard deviation). For the three real-world datasets, all estimators are evaluated against the reference causal effect values in the literature. 

\paragraph{Implementation Details.} We use \textit{Python} and the libraries including \textit{pytorch}~\cite{paszke2019pytorch}, \textit{pyro}~\cite{bingham2019pyro} and \textit{scikit-learn}~\cite{pedregosa2011scikit} to implement our CIV.VAE method. We provide the details of our CIV.VAE implementation and the parameters setting in the supplement. TSLS is implemented by using the functions \textit{glm} and \textit{ivglm} in the \textit{R} packages \textit{stats} and \textit{ivtools}~\cite{sjolander2019instrumental} respectively. FIVR is coded by employing the function \emph{instrumental}$\_$\emph{forest} in the \textit{R} package \emph{grf}~\cite{athey2019generalized}. The program of DeepIV is retrieved from the authors' GitHub~\footnote{\url{https://github.com/jhartford/DeepIV}}. The implementations of OrthoIV and DMLIV are from the \textit{Python} package \textit{encoml}. IV.Tetrad is obtained from the authors' site~\footnote{\url{http://www.homepages.ucl.ac.uk/~ucgtrbd/code/iv_discovery}}. The implementation of CEVAE is obtained from the \textit{Python} library \textit{pyro}~\cite{bingham2019pyro} and the implementation of TEDVAE is downloaded from the authors' GitHub~\footnote{\url{https://github.com/WeijiaZhang24/TEDVAE}}.

\begin{table*}[t]
	\centering
	\begin{tabular}{|cccccccc|}
		\hline
		\multicolumn{2}{|c|}{}   & \multicolumn{6}{|c|}{Samples} \\ \cline{3-8} 
		\multicolumn{2}{|c|}{Methods}    & \multicolumn{1}{c|}{2k}              & \multicolumn{1}{c|}{4k}              & \multicolumn{1}{c|}{6k}    & \multicolumn{1}{c|}{8k}    & \multicolumn{1}{c|}{10k}   & 20k   \\ \hline
		\multicolumn{1}{|c|}{\multirow{5}{*}{Known IV}}   & \multicolumn{1}{c|}{TSLS}      & \multicolumn{1}{c|}{11.12$\pm$1.74}           & \multicolumn{1}{c|}{11.12$\pm$1.38}           & \multicolumn{1}{c|}{10.87$\pm$0.83} & \multicolumn{1}{c|}{11.08$\pm$0.94} & \multicolumn{1}{c|}{11.34$\pm$0.82} & 11.06$\pm$0.55\\ \cline{2-8} 
		\multicolumn{1}{|c|}{}                     & \multicolumn{1}{c|}{FIVR}    & \multicolumn{1}{c|}{1.58$\pm$0.97} & \multicolumn{1}{c|}{1.10$\pm$0.61} & \multicolumn{1}{c|}{0.64$\pm$0.50} & \multicolumn{1}{c|}{0.62$\pm$0.37} & \multicolumn{1}{c|}{0.63$\pm$0.42} & 0.33$\pm$0.21\\ \cline{2-8} 
		\multicolumn{1}{|c|}{}                     & \multicolumn{1}{c|}{DeepIV}    & \multicolumn{1}{c|}{1.64$\pm$0.19} & \multicolumn{1}{c|}{1.47$\pm$0.21} & \multicolumn{1}{c|}{1.53$\pm$0.21} & \multicolumn{1}{c|}{1.53$\pm$0.23} & \multicolumn{1}{c|}{1.43$\pm$0.32} & 1.33$\pm$0.22\\ \cline{2-8} 
		\multicolumn{1}{|c|}{}  & \multicolumn{1}{c|}{OrthIV}    & \multicolumn{1}{c|}{3.57$\pm$2.90}     & \multicolumn{1}{c|}{1.91$\pm$1.19}    & \multicolumn{1}{c|}{1.29$\pm$1.49} & \multicolumn{1}{c|}{1.51$\pm$1.25} & \multicolumn{1}{c|}{1.29$\pm$0.93} & 0.71$\pm$0.52 \\ \cline{2-8} 
		\multicolumn{1}{|c|}{}                     & \multicolumn{1}{c|}{DMLIV}     & \multicolumn{1}{c|}{3.53$\pm$2.63}           & \multicolumn{1}{c|}{2.11$\pm$1.70}           & \multicolumn{1}{c|}{1.19$\pm$1.33} & \multicolumn{1}{c|}{1.49$\pm$1.32} & \multicolumn{1}{c|}{1.12$\pm$0.85} & 0.71$\pm$0.58 \\ \hline
		\multicolumn{1}{|c|}{\multirow{2}{*}{Unknown IV}}  & \multicolumn{1}{c|}{sisVIVE}   & \multicolumn{1}{c|}{1.37$\pm$0.79} & \multicolumn{1}{c|}{1.61$\pm$0.97} & \multicolumn{1}{c|}{1.70$\pm$1.01} & \multicolumn{1}{c|}{1.39$\pm$0.61} & \multicolumn{1}{c|}{1.56$\pm$0.81} & 2.08$\pm$0.89\\ \cline{2-8}
		\multicolumn{1}{|c|}{} & \multicolumn{1}{c|}{IV.Tetrad} & \multicolumn{1}{c|}{2.89$\pm$3.77} &\multicolumn{1}{c|}{2.15$\pm$3.45}     & \multicolumn{1}{c|}{2.73$\pm$3.92} & \multicolumn{1}{c|}{2.86$\pm$3.90} & \multicolumn{1}{c|}{2.12$\pm$3.60} & 2.90$\pm$3.98 \\ \hline
		\multicolumn{1}{|c|}{\multirow{2}{*}{VAE-based}} & \multicolumn{1}{c|}{CEVAE} & \multicolumn{1}{c|}{\textbf{1.34$\pm$0.23}} & \multicolumn{1}{c|}{1.14$\pm$0.25}     & \multicolumn{1}{c|}{1.23$\pm$0.29} & \multicolumn{1}{c|}{1.17$\pm$0.36} & \multicolumn{1}{c|}{1.17$\pm$0.39} & 1.13$\pm$0.58 \\ \cline{2-8} 
		\multicolumn{1}{|c|}{}                     & \multicolumn{1}{c|}{TEDVAE}    & \multicolumn{1}{c|}{1.68$\pm$0.27}           & \multicolumn{1}{c|}{1.65$\pm$0.18}           & \multicolumn{1}{c|}{1.70$\pm$0.14} & \multicolumn{1}{c|}{1.68$\pm$0.11} & \multicolumn{1}{c|}{1.70$\pm$0.10} & 1.70$\pm$0.09 \\ \hline
		\multicolumn{2}{|c|}{CIV.VAE}    & \multicolumn{1}{c|}{1.94$\pm$1.45}           & \multicolumn{1}{c|}{\textbf{0.95$\pm$0.42}}           & \multicolumn{1}{c|}{\textbf{0.42$\pm$0.34}} & \multicolumn{1}{c|}{\textbf{0.33$\pm$0.20}} & \multicolumn{1}{c|}{\textbf{0.26$\pm$0.19}} & \textbf{0.22$\pm$0.17} \\ \hline
	\end{tabular}
	\caption{The table summarises the estimated errors $\varepsilon_{ACE}$ (Mean$\pm$STD) over 30 synthetic datasets in each sample size. The lowest estimated errors are marked in boldface. Note that CIV.VAE relies on the least domain knowledge among all estimators and obtain the smallest $\varepsilon_{ACE}$ among all methods compared.  } 
	\label{tab:syn_results}
\end{table*} 

\subsection{Evaluation with Simulated Data}
\label{subsec:simu}

We use the causal DAG provided in Fig. 1 in the supplement to generate a set of synthetic datasets with a range of sample sizes: 2k, 4k, 6k, 8k, 10k and 20k. The set of $\mathbf{X}$ is $\{S, X_1, X_2, X_3, X_4, X_5\}$ and the set $\mathbf{U}$ (latent confounders) consists of $\{U, U_1, U_2, U_3, U_4\}$ where $T\leftarrow U \rightarrow Y$ and $\mathbf{U}'= \{U_1, U_2, U_3, U_4\}$. The true $ACE(T, Y)$ of all synthetic datasets is 2. Due to page limitation, more details of the data generation process are provided in the supplement. 

To avoid the bias brought by data generation, 30 synthetic datasets for each sample size are generated in our experiments. We utilise the CIV $S$ in the underlying causal DAG as a standard IV for TSLS. Moreover, the CIV $S$ and the set $\mathbf{X}\setminus\{S\}$ in the underlying causal DAG are the true CIV and the corresponding conditioning set, respectively, and they both are taken as input for the four estimators, FIVR, DeepIV, OrthIV and DMLIV. The estimation errors of all estimators on all synthetic datasets are reported in Table~\ref{tab:syn_results}.

\paragraph{Results.} From Table~\ref{tab:syn_results}, we see that CIV.VAE obtains the smallest $\varepsilon_{ACE}$ across almost all datasets compared with the other estimators. Note that CIV.VAE relies on the least domain knowledge, i.e. does not require a specific IV (whereas TSLS, FIVR, DeepIV, OrthIV and DMLIV do) or the conditioning set (whereas IV.Tetrad does), or a rich set of IVs (whereas sisVIVE does) or a rich set of proxy variables (whereas CEVAE does) or the unconfoundedness assumption (whereas TEDVAE does).  

There are five other observations from Table~\ref{tab:syn_results}: (1) the baseline IV estimator, TSLS, has the largest estimation errors because the confounding bias between $S$ and $Y$ caused by confounders and latent confounders is not blocked at all even though it uses the CIV $S$ as its known IV. (2) FIVR obtains the best performance in the first type of IV-based methods, \ie TSLS, DeepIV, OrthIV and DMLIV, but has larger estimation errors than CIV.VAE. (3) two IV-based estimators without needing a given IV, sisVIVE and IV.Tetrad obtain constant estimation errors with little variation across all synthetic datasets and have larger estimation errors than CIV.VAE. (4) the two VAE-based causal effect estimators in the third type of comparison methods, CEVAE and TEDVAE obtain smaller estimation errors than the second type of methods, but both have larger estimation errors than CIV.VAE. (5) As the sample size increases, CIV.VAE consistently gets the lowest bias compared with all causal effect estimators except for the 2k sample size. This indicates that CIV.VAE requires relatively large sample size to learn the two representations $\mathbf{Z}_{T}$ and $\mathbf{Z}_{\mathbf{C}}$ such that their distributions are close to the true distributions. 

Therefore, the experimental results on synthetic datasets show that CIV.VAE has the capability to learn high quality CIV and conditioning set representations for causal effect estimation from data with latent confounders.

\subsection{Experiments on Real-world Datasets}
\label{Subsec:realworlddatasets}
In this section, we conduct experiments on three benchmark real-world datasets, Schoolingreturns~\cite{card1993using}, 401(k)~\cite{verbeek2008guide} and Sachs~\cite{sachs2005causal} for which the empirical causal effects available and widely accepted. Note that the first two datasets each have a known IV based on domain knowledge, but Sachs does not have a known IV. The detailed descriptions of the three real-world datasets are introduced in the supplement.  

\begin{table*}[t]
	\centering
	\small  
	\begin{tabular}{|c|c|c|c|c|c|c|c|c|c|c|c|}
		\hline
		Estimators       &  TSLS  & FIVR   & DeepIV  &OrthIV & DMLIV & sisVIVE & IV.Tetrad & CEVAE & TEDVAE & CIV.VAE \\ \hline
		Schoolingreturns  &   0.5042 & 1.1513 & -0.0444 & 1.3189 & 1.2806 & 0.0254 & \textbf{0.0643} & \textbf{0.0956}& -0.1082 & \textbf{0.1034}  \\ \hline
		401(k) &  0.1500 & \textbf{0.0746}& - & 0.1502 & 0.1503 & 1.5172 &  1.2484 & 0.0384 & 0.0283 & \textbf{0.0752}\\ \hline
		Sachs  &  - & - & - & - &- &  \textbf{0.4356} & \textbf{1.4301}  & \textbf{0.2542} & \textbf{0.2553} & \textbf{1.5133 }     \\ \hline
	\end{tabular}
	\caption{The estimated causal effects of all estimators on the three real-world datasets. We highlight the estimated causal effects within the 95\% confidence interval on Schoolingreturns and 401(k). `-' is used for the corresponding IV-based estimator on Sachs because there is not a known IV. Note that DeepIV cannot work on 401(k) and is also marked as `-'. }
	\label{tab:results}
\end{table*}

\paragraph{Schoolingreturns.} This dataset consists of 3,010 records and 19 variables. The treatment variable is the education level of a person. The outcome variable is raw wages in 1976 (in cents per hour). The goal of collecting this dataset is to study the causal effect of the education level on wages. In the work~\cite{card1993using}, \emph{nearcollege} (geographical proximity to a college) is nominated as the known IV, and the estimated $ACE(T, Y) = 0.1329$ with 95\% confidence interval (0.0484, 0.2175) from the works~\cite{verbeek2008guide} as the reference causal effect.  

\paragraph{401(k).} The dataset contains 9,275 individuals and 11 variables. The dataset is from the survey of income and program participation (SIPP)~\cite{verbeek2008guide}.  The treatment is \emph{p401k} (a binary indicated variable of participation in 401 (k)), and the outcome is \emph{pira} (a binary indicated variable, $pira$ = 1 denotes participation in IRA). \emph{e401k} (a binary indicated  variable of eligibility for 401 (k)) is used as an IV,  $ACE(T, Y)= 0.0712$  with 95\% confidence interval (0.047, 0.095)~\cite{verbeek2008guide} as the reference causal effect.

\paragraph{Sachs.} The dataset contains 853 samples and 11 variables~\cite{sachs2005causal}. The treatment is $Erk$ (the manipulation of concentration levels of a molecule). The outcome is the concentration of $Akt$. Note that there is not a nominated IV. In this work, we take the reported $ACE(T, Y) =1.4301$  with 95\% confidence interval (0.05, 3.23) in the work~\cite{silva2017learning} (\ie IV.Tetrad's estimated causal effect) as the reference causal effect.

\paragraph{Results.}  All results on the three datasets are reported in Table~\ref{tab:results}. From the results in Table~\ref{tab:results}, we see that (1) the causal effects estimated by CIV.VAE are within the 95\% confidence interval of their empirical results, and on Sachs, the estimated causal effect by CIV.VAE is the closest to IV.Tetrad's result; (2) the estimated causal effects by IV.Tetrad and CEVAE on Schoolingreturns, and the estimation by FIVR on 401(k) are in the 95\% empirical interval, but not on all three datasets. The other compared estimators do not perform well on both datasets with empirical intervals. (3) There is not a known IV on Sachs, so IV-based estimators requiring a given IV do not work on this dataset. Note that the estimated average causal effects by CIV.VAE, sisVIVE, CEVAE and TEDVAE are in the empirical interval and they work well on Sachs.

The experiments on the three real-world datasets further confirm that CIV.VAE is able to infer the conditional IV representation $\mathbf{Z}_{T}$ and the corresponding conditioning representation $\mathbf{Z}_{\mathbf{C}}$ from data in the presence of latent confounders for unbiased average causal effect estimation. 

In a word, \textit{CIV.VAE}, without knowing a CIV and the corresponding conditioning set, performs better than the state-of-the-art IV-based and two VAE-based estimators on the three real-world datasets.

\paragraph{Limitations.} CIV.VAE relies on the assumptions of the pretreatment variables and the existence of at least one CIV, and it also relies on that VAE correctly identifies the latent variables. However, work by~\cite{khemakhem2020variational} shows that identifying latent variables is not guaranteed by VAE. This means that, when some of the assumptions are not satisfied or VAE identifiability, CIV.VAE may provide an unreliable conclusion. To avoid the potential negative impact, it is better to choose other causal effect estimators to cross check or conducting a sensitive analysis~\cite{imbens2015causal,hartford2021valid}. Furthermore, the identifiable VAE framework (a.k.a. iVAE) in~\cite{khemakhem2020variational} provides some ideas for improving the identifiability of the VAE-based model, and it can be used to improve the reliability of CIV.VAE.

\section{Related Work}
\label{sec:relwork}	
We review the work closely related to our proposed method, including IV-based methods requiring a given IV and data-driven IV estimators without a known IV.

\paragraph{IV-based estimators requiring a given IV.}  Several IV-based causal effect estimators have been proposed for average causal effect estimation when there is a known IV, such as causal random forest based IV regression (FIVR)~\cite{athey2019generalized}, generalised method of moments based IV estimator (GMM)~\cite{bennett2019deep}, deep ensemble method based IV approach (DeepIV)~\cite{hartford2017deep} and kernel IV regression (KIV)~\cite{singh2019kernel}. Different from these IV-based estimators, CIV.VAE does not require a given IV and the conditioning set by domain knowledge. 

\paragraph{Data-driven IV-based estimators without a known IV.} In most real-world applications, there is not a known IV. A few data-driven IV estimators have been developed for discovering a valid IV~\cite{yuan2022auto} or a synthesising IV~\cite{burgess2013use} or eliminating the influence of invalid IVs by using statistical strategies~\cite{kang2016instrumental,guo2018confidence,hartford2021valid}. For instance, the tetrad constraint is utilised by IV.Tetrad~\cite{silva2017learning} to validate the validity of a pair of CIVs for estimating causal effects from data in presence of latent confounders. However, it requires that there exists at least a pair of CIVs and assumes that the set of all the remained variables is the conditioning set. Kuang \etal~\cite{kuang2020ivy} proposed the Ivy method to synthesise an IV by combining a set of IV candidates for determining all invalid IVs or dependencies. The sisVIVE method~\cite{kang2016instrumental} was developed to estimate causal effects when the majority assumption holds (\ie at least a half of the covariates are valid IVs). Under the majority assumption, the ModeIV estimator~\cite{hartford2021valid} is developed by employing a deep learning based IV estimator~\cite{hartford2017deep}. Unlike this type of data-driven IV-based estimators, CIV.VAE takes the advantages of deep generative model to learn the latent IV representation and the conditioning set representation from data.

\section{Conclusion}
\label{sec:con}
Latent confounders are a crucial challenge for causal inference in practice. IV-based approach provides an effective way to circumvent the latent confounding problem. However, for data-driven causal inference, standard IV is not feasible due to the strict conditions. CIVs shed light on data-driven IV-based causal inference, but in many cases it is impossible to distinguish a CIV and its conditioning set from data using traditional methods. In this paper, by leveraging the VAE model, we have designed the CIV.VAE method to learn the representations of a CIV and its conditioning set from data with latent confounders. We have conducted extensive experiments on synthetic and three real-world datasets, and the experimental results demonstrate the capability and the validity of our proposed CIV.VAE model against the state-of-the-art estimators in causal effect estimation with data containing latent confounders.

\bibliographystyle{AAAI}

\bibliography{aaai23}

\begin{thebibliography}{44}
\providecommand{\natexlab}[1]{#1}

\bibitem[{Abadie(2003)}]{abadie2003semiparametric}
Abadie, A. 2003.
\newblock Semiparametric instrumental variable estimation of treatment response
  models.
\newblock \emph{Journal of econometrics}, 113(2): 231--263.

\bibitem[{Angrist and Imbens(1995)}]{angrist1995two}
Angrist, J.~D.; and Imbens, G.~W. 1995.
\newblock Two-stage least squares estimation of average causal effects in
  models with variable treatment intensity.
\newblock \emph{Journal of the American statistical Association}, 90(430):
  431--442.

\bibitem[{Athey, Tibshirani, and Wager(2019)}]{athey2019generalized}
Athey, S.; Tibshirani, J.; and Wager, S. 2019.
\newblock Generalized random forests.
\newblock \emph{The Annals of Statistics}, 47(2): 1148--1178.

\bibitem[{Bennett, Kallus, and Schnabel(2019)}]{bennett2019deep}
Bennett, A.; Kallus, N.; and Schnabel, T. 2019.
\newblock Deep generalized method of moments for instrumental variable
  analysis.
\newblock In \emph{International Conference on Neural Information Processing
  Systems}, 3564--3574.

\bibitem[{Bingham, Chen et~al.(2019)}]{bingham2019pyro}
Bingham, E.; Chen, J.~P.; et~al. 2019.
\newblock Pyro: {D}eep universal probabilistic programming.
\newblock \emph{The Journal of Machine Learning Research}, 20(1): 973--978.

\bibitem[{Brito and Pearl(2002)}]{brito2002generalized}
Brito, C.; and Pearl, J. 2002.
\newblock Generalized instrumental variables.
\newblock In \emph{Proceedings of the Eighteenth conference on Uncertainty in
  artificial intelligence}, 85--93.

\bibitem[{Burgess and Thompson(2013)}]{burgess2013use}
Burgess, S.; and Thompson, S.~G. 2013.
\newblock Use of allele scores as instrumental variables for {M}endelian
  randomization.
\newblock \emph{International Journal of Epidemiology}, 42(4): 1134--1144.

\bibitem[{Card(1993)}]{card1993using}
Card, D. 1993.
\newblock Using geographic variation in college proximity to estimate the
  return to schooling.

\bibitem[{Chernozhukov et~al.(2018)Chernozhukov, Chetverikov, Demirer, Duflo,
  Hansen, Newey, and Robins}]{chernozhukov2018double}
Chernozhukov, V.; Chetverikov, D.; Demirer, M.; Duflo, E.; Hansen, C.; Newey,
  W.; and Robins, J. 2018.
\newblock Double/debiased machine learning for treatment and structural
  parameters.
\newblock \emph{The Econometrics Journal}, 21(1): C1--C68.

\bibitem[{Guo et~al.(2020)Guo, Cheng, Li, Hahn, and Liu}]{guo2020survey}
Guo, R.; Cheng, L.; Li, J.; Hahn, P.~R.; and Liu, H. 2020.
\newblock A survey of learning causality with data: Problems and methods.
\newblock \emph{ACM Computing Surveys (CSUR)}, 53(4): 1--37.

\bibitem[{Guo, Kang et~al.(2018)}]{guo2018confidence}
Guo, Z.; Kang, H.; et~al. 2018.
\newblock Confidence intervals for causal effects with invalid instruments by
  using two-stage hard thresholding with voting.
\newblock \emph{Journal of the Royal Statistical Society: Series B (Statistical
  Methodology)}, 80(4): 793--815.

\bibitem[{Hartford, Lewis et~al.(2017)}]{hartford2017deep}
Hartford, J.; Lewis, G.; et~al. 2017.
\newblock Deep {IV}: {A} flexible approach for counterfactual prediction.
\newblock In \emph{International Conference on Machine Learning}, 1414--1423.

\bibitem[{Hartford et~al.(2021)Hartford, Veitch, Sridhar, and
  Leyton-Brown}]{hartford2021valid}
Hartford, J.~S.; Veitch, V.; Sridhar, D.; and Leyton-Brown, K. 2021.
\newblock Valid causal inference with (some) invalid instruments.
\newblock In \emph{International Conference on Machine Learning}, 4096--4106.
  PMLR.

\bibitem[{Hassanpour and Greiner(2019)}]{hassanpour2019learning}
Hassanpour, N.; and Greiner, R. 2019.
\newblock Learning disentangled representations for counterfactual regression.
\newblock In \emph{International Conference on Learning Representations},
  1--11.

\bibitem[{Hern{\'a}n and Robins(2006)}]{hernan2006instruments}
Hern{\'a}n, M.~A.; and Robins, J.~M. 2006.
\newblock Instruments for causal inference: an epidemiologist's dream?
\newblock \emph{Epidemiology}, 360--372.

\bibitem[{Imbens and Rubin(2015)}]{imbens2015causal}
Imbens, G.~W.; and Rubin, D.~B. 2015.
\newblock \emph{Causal {I}nference in {S}tatistics, {S}ocial, and {B}iomedical
  {S}ciences}.
\newblock Cambridge University Press.

\bibitem[{Kang et~al.(2016)Kang, Zhang, Cai, and Small}]{kang2016instrumental}
Kang, H.; Zhang, A.; Cai, T.~T.; and Small, D.~S. 2016.
\newblock Instrumental variables estimation with some invalid instruments and
  its application to Mendelian randomization.
\newblock \emph{Journal of the American statistical Association}, 111(513):
  132--144.

\bibitem[{Khemakhem et~al.(2020)Khemakhem, Kingma, Monti, and
  Hyvarinen}]{khemakhem2020variational}
Khemakhem, I.; Kingma, D.; Monti, R.; and Hyvarinen, A. 2020.
\newblock Variational autoencoders and nonlinear ica: A unifying framework.
\newblock In \emph{International Conference on Artificial Intelligence and
  Statistics}, 2207--2217. PMLR.

\bibitem[{Kingma and Welling(2014)}]{kingma2014auto}
Kingma, D.~P.; and Welling, M. 2014.
\newblock Auto-encoding variational bayes.
\newblock In \emph{International Conference on Learning Representations}.

\bibitem[{Kingma, Welling et~al.(2019)}]{kingma2019introduction}
Kingma, D.~P.; Welling, M.; et~al. 2019.
\newblock An introduction to variational autoencoders.
\newblock \emph{Foundations and Trends{\textregistered} in Machine Learning},
  12(4): 307--392.

\bibitem[{Kuang, Sala et~al.(2020)}]{kuang2020ivy}
Kuang, Z.; Sala, F.; et~al. 2020.
\newblock Ivy: {I}nstrumental variable synthesis for causal inference.
\newblock In \emph{International Conference on Artificial Intelligence and
  Statistics}, 398--410.

\bibitem[{Kuroki and Pearl(2014)}]{kuroki2014measurement}
Kuroki, M.; and Pearl, J. 2014.
\newblock Measurement bias and effect restoration in causal inference.
\newblock \emph{Biometrika}, 101(2): 423--437.

\bibitem[{Louizos et~al.(2017)Louizos, Shalit, Mooij, Sontag, Zemel, and
  Welling}]{louizos2017causal}
Louizos, C.; Shalit, U.; Mooij, J.~M.; Sontag, D.; Zemel, R.; and Welling, M.
  2017.
\newblock Causal effect inference with deep latent-variable models.
\newblock In \emph{Advances in Neural Information Processing Systems},
  6446--6456.

\bibitem[{Miao, Geng, and Tchetgen~Tchetgen(2018)}]{miao2018identifying}
Miao, W.; Geng, Z.; and Tchetgen~Tchetgen, E.~J. 2018.
\newblock Identifying causal effects with proxy variables of an unmeasured
  confounder.
\newblock \emph{Biometrika}, 105(4): 987--993.

\bibitem[{Paszke, Gross et~al.(2019)}]{paszke2019pytorch}
Paszke, A.; Gross, S.; et~al. 2019.
\newblock PyTorch: an imperative style, high-performance deep learning library.
\newblock In \emph{International Conference on Neural Information Processing
  Systems}, 8026--8037.

\bibitem[{Pearl(2009)}]{pearl2009causality}
Pearl, J. 2009.
\newblock \emph{Causality}.
\newblock Cambridge university press.

\bibitem[{Pearl and Mackenzie(2018)}]{pearl2018book}
Pearl, J.; and Mackenzie, D. 2018.
\newblock \emph{The {B}ook of {W}hy: the {N}ew {S}cience of {C}ause and
  {E}ffect}.
\newblock Basic Books.

\bibitem[{Pedregosa et~al.(2011)Pedregosa, Varoquaux, Gramfort
  et~al.}]{pedregosa2011scikit}
Pedregosa, F.; Varoquaux, G.; Gramfort, A.; et~al. 2011.
\newblock Scikit-learn: Machine learning in Python.
\newblock \emph{the Journal of machine Learning research}, 12: 2825--2830.

\bibitem[{Rissanen and Marttinen(2021)}]{rissanen2021critical}
Rissanen, S.; and Marttinen, P. 2021.
\newblock A critical look at the consistency of causal estimation with deep
  latent variable models.
\newblock In \emph{International Conference on Neural Information Processing
  Systems}, 4207--4217.

\bibitem[{Sachs, Perez et~al.(2005)}]{sachs2005causal}
Sachs, K.; Perez, O.; et~al. 2005.
\newblock Causal protein-signaling networks derived from multiparameter
  single-cell data.
\newblock \emph{Science}, 308(5721): 523--529.

\bibitem[{Sch{\"o}lkopf(2022)}]{scholkopf2022causality}
Sch{\"o}lkopf, B. 2022.
\newblock Causality for machine learning.
\newblock In \emph{Probabilistic and Causal Inference: The Works of Judea
  Pearl}, 765--804.

\bibitem[{Sch{\"o}lkopf et~al.(2021)Sch{\"o}lkopf, Locatello, Bauer, Ke,
  Kalchbrenner, Goyal, and Bengio}]{scholkopf2021toward}
Sch{\"o}lkopf, B.; Locatello, F.; Bauer, S.; Ke, N.~R.; Kalchbrenner, N.;
  Goyal, A.; and Bengio, Y. 2021.
\newblock Toward causal representation learning.
\newblock \emph{Proceedings of the IEEE}, 109(5): 612--634.

\bibitem[{Silva and Shimizu(2017)}]{silva2017learning}
Silva, R.; and Shimizu, S. 2017.
\newblock Learning instrumental variables with structural and non-gaussianity
  assumptions.
\newblock \emph{Journal of Machine Learning Research}, 18(120): 1--49.

\bibitem[{Singh, Sahani, and Gretton(2019)}]{singh2019kernel}
Singh, R.; Sahani, M.; and Gretton, A. 2019.
\newblock Kernel instrumental variable regression.
\newblock In \emph{International Conference on Neural Information Processing
  Systems}, 4593--4605.

\bibitem[{Sjolander and Martinussen(2019)}]{sjolander2019instrumental}
Sjolander, A.; and Martinussen, T. 2019.
\newblock Instrumental variable estimation with the R package ivtools.
\newblock \emph{Epidemiologic Methods}, 8(1).

\bibitem[{Sohn, Yan, and Lee(2015)}]{sohn2015learning}
Sohn, K.; Yan, X.; and Lee, H. 2015.
\newblock Learning structured output representation using deep conditional
  generative models.
\newblock In \emph{Proceedings of the 28th International Conference on Neural
  Information Processing Systems-Volume 2}, 3483--3491.

\bibitem[{Spirtes et~al.(2000)Spirtes, Glymour, Scheines, and
  Heckerman}]{spirtes2000causation}
Spirtes, P.; Glymour, C.~N.; Scheines, R.; and Heckerman, D. 2000.
\newblock \emph{Causation, prediction, and search}.
\newblock MIT press.

\bibitem[{Syrgkanis, Lei et~al.(2019)}]{syrgkanis2019machine}
Syrgkanis, V.; Lei, V.; et~al. 2019.
\newblock Machine learning estimation of heterogeneous treatment effects with
  instruments.
\newblock In \emph{International Conference on Neural Information Processing
  Systems}, 15193--15202.

\bibitem[{Van~der Zander, Li{\'s}kiewicz, and
  Textor(2015)}]{van2015efficiently}
Van~der Zander, B.; Li{\'s}kiewicz, M.; and Textor, J. 2015.
\newblock Efficiently finding conditional instruments for causal inference.
\newblock 3243--3249.

\bibitem[{Verbeek(2008)}]{verbeek2008guide}
Verbeek, M. 2008.
\newblock \emph{A guide to modern econometrics}.
\newblock John Wiley \& Sons.

\bibitem[{Wooldridge(2010)}]{wooldridge2010econometric}
Wooldridge, J.~M. 2010.
\newblock \emph{Econometric analysis of cross section and panel data}.
\newblock MIT press.

\bibitem[{Yao et~al.(2021)Yao, Chu, Li, Li, Gao, and Zhang}]{yao2021survey}
Yao, L.; Chu, Z.; Li, S.; Li, Y.; Gao, J.; and Zhang, A. 2021.
\newblock A survey on causal inference.
\newblock \emph{ACM Transactions on Knowledge Discovery from Data (TKDD)},
  15(5): 1--46.

\bibitem[{Yuan, Wu et~al.(2022)}]{yuan2022auto}
Yuan, J.; Wu, A.; et~al. 2022.
\newblock Auto {IV}: Counterfactual Prediction via Automatic Instrumental
  Variable Decomposition.
\newblock \emph{ACM Transactions on Knowledge Discovery from Data}, 16(4):
  1--20.

\bibitem[{Zhang, Liu, and Li(2021)}]{zhang2021treatment}
Zhang, W.; Liu, L.; and Li, J. 2021.
\newblock Treatment Effect Estimation with Disentangled Latent Factors.
\newblock In \emph{The AAAI Conference on Artificial Intelligence},
  10923--10930.

\end{thebibliography}

\vspace{20pt}

\appendix

\section{Supplement}
The supplement is provided to the paper `Causal Inference with Conditional Instruments using Deep Generative Models'. We introduce some definitions of graphical causal modelling and details of simulated and real-world datasets.
 
 \section{Preliminary}
 \label{app:sec:Preliminaries}
 \subsection{Notations \& Definitions}
 In a causal DAG $\mathcal{G}=(\mathbf{V}, \mathbf{E})$, if a path $\pi$ contains $V_k \astrightarrow V_i \astleftarrow V_l$ as a subpath, then $V_{i}$ is a collider. A \emph{collider path} in the DAG $\mathcal{G}$ is a path on which every non-endpoint node is a collider. A path of length one is a \emph{trivial collider path}.

 The definitions of Markov property and faithfulness are introduced in the following.
 \begin{definition}[Markov property~\cite{pearl2009causality}]
 	\label{Markov condition}
 	Given a DAG $\mathcal{G}=(\mathbf{V}, \mathbf{E})$ and the joint probability distribution of $\mathbf{V}$ $(P(\mathbf{V}))$, $\mathcal{G}$ satisfies the Markov property if for $\forall V_i \in \mathbf{V}$, $V_i$ is probabilistically independent of all of its non-descendants, given the parent nodes of $V_i$.
 \end{definition}
 
 \begin{definition}[Faithfulness~\cite{spirtes2000causation}]
 	\label{Faithfulness}
 	A DAG $\mathcal{G}=(\mathbf{V}, \mathbf{E})$ is faithful to a joint distribution $P(\mathbf{V})$ over the set of variables $\mathbf{V}$ if and only if every independence present in $P(\mathbf{V})$ is entailed by $\mathcal{G}$ and satisfies the Markov property. A joint distribution $P(\mathbf{V})$ over the set of variables $\mathbf{V}$ is faithful to the DAG $\mathcal{G}$ if and only if the DAG $\mathcal{G}$ is faithful to the joint distribution $P(\mathbf{V})$.
 \end{definition}
 
 When the faithfulness assumption is satisfied regarding a joint distribution $P(\mathbf{V})$ and a DAG of a set of variables $\mathbf{V}$, the dependency/independency relations among the variables can be read from the DAG. In a DAG, d-separation is a graphical criterion that is used to read off the identification of conditional independence between variables entailed in the DAG when the Markov property and faithfulness are satisfied~\cite{pearl2009causality,spirtes2000causation}.
 
 \begin{definition}[d-separation~\cite{pearl2009causality}]
 	\label{d-separation}
 	A path $\pi$ in a DAG $\mathcal{G}=(\mathbf{V}, \mathbf{E})$ is said to be d-separated (or blocked) by a set of nodes $\mathbf{W}$ if and only if
 	(i) $\pi$ contains a chain $V_i \rightarrow V_k \rightarrow V_j$ or a fork $V_i \leftarrow V_k \rightarrow V_j$ such that the middle node $V_k$ is in $\mathbf{W}$, or
 	(ii) $\pi$ contains a collider $V_k$ such that $V_k$ is not in $\mathbf{W}$ and no descendant of $V_k$ is in $\mathbf{W}$.
 	A set $\mathbf{W}$ is said to d-separate $V_i$ from $V_j$ ($ V_i \indep V_j\mid\mathbf{W}$) if and only if $\mathbf{W}$ blocks every path between $V_i$ to $V_j$. 
 	Otherwise, they are said to be d-connected by $\mathbf{W}$, denoted as $V_i\nindep V_j\mid\mathbf{W}$.
 \end{definition}

\section{Experiments}
\subsection{Reproducibility}
In this section, we provide the parameter settings of our CIV.VAE for reproducibility purpose. Our CIV.VAE were implemented by \textit{Python} and the libraries \textit{pytorch}~\cite{paszke2019pytorch} and \textit{pyro}~\cite{bingham2019pyro}. 
 The details of the parameter settings of our CIV.VAE for simulated datasets, Schoolreturning, 401(k) and Sachs are reported in Table~\ref{tab:001}. The major parameters are described as follows:

\begin{description}
	\item [latent\_dim\_Zt ($|\mathbf{Z}_T|$):] the dimensions of the latent IV representation $\mathbf{Z}_T$
	\item [latent\_dim\_Zc ($|\mathbf{Z}_\mathbf{C}|$):] the dimensions of the latent representation $\mathbf{Z}_\mathbf{C}$ of the conditioning set
	\item [num\_epochs (epoch):] one Epoch is when an entire dataset is passed forward and backward through the neural network only once
	\item [hidden\_dim ($ls$):] the number of hidden layers
	\item [batch\_size ($bs$):] total number of training examples present in a single batch 
	\item [learning\_rate ($lr$):] the learning rate
\end{description}

\begin{table}[t]
	\centering
	\footnotesize  
	\begin{tabular}{|c|c|c|c|c|}
		\hline
		Parameter & Simulation & Schoolingreturn & 401(k) & Sachs \\ \hline
		$|\mathbf{Z}_T|$ &      1              &          1       &    1    &    1   \\ \hline
	$|\mathbf{Z}_\mathbf{C}|$	&     3               &   2              &      2  &   2    \\ \hline
	epoch	&         100       &        100         &     100   &    100   \\ \hline
	$ls$	&        200            &     200            &   200     &  200     \\ \hline
	$bs$	&     256               &     256            &   256     &     256  \\ \hline
	$lr$	&    0.001                &         0.001         &      0.001   &   0.001     \\ \hline
	\end{tabular}
	\caption{Details of the parameter settings in CIV.VAE method for both simulated data and real-world datasets.}
	\label{tab:001}
\end{table}

\subsection{Evaluation with Simulated Data}
 \label{subsec:simulation}
\begin{figure}[ht]
	\centering
	\includegraphics[scale=0.36]{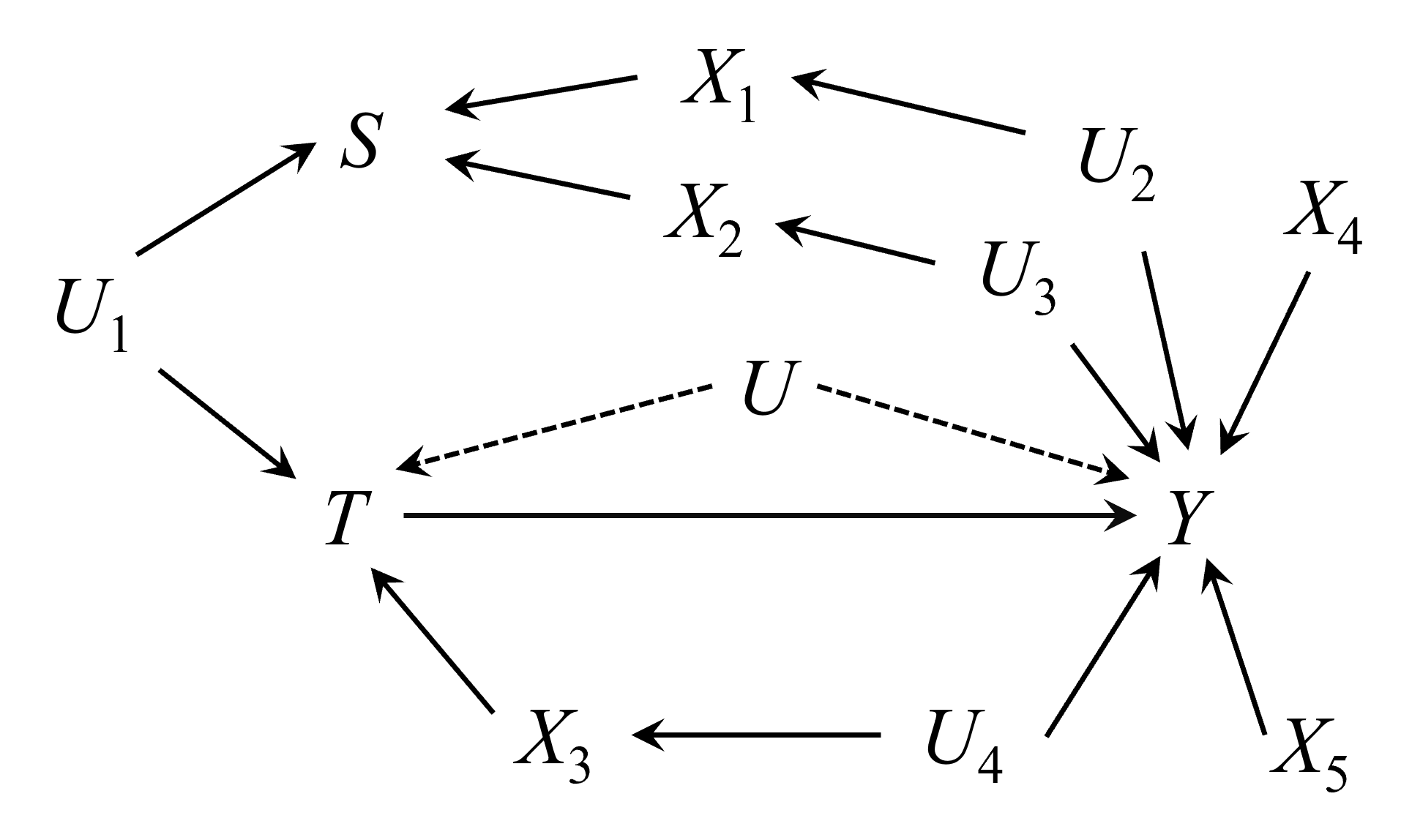}
	\caption{The true causal DAG with a latent confounder $U$ between $T$ and $Y$ is used to generate the simulated datasets. $\mathbf{U}_\mathbf{C}= \{U\}$,  $\mathbf{U}'= \{U_1, U_2, U_3, U_4\}$ are four latent confounders, and $\mathbf{X}= \{S, X_1, X_2, X_3, X_4, X_5\}$ are pretreatment variables. Note that $S$ is a CIV conditioning on $\{X_1, X_2\}$. }
	\label{fig:figure_001}
\end{figure}

The true DAG over $\mathbf{X}\cup\mathbf{U}\cup\{T, Y\}$ as shown in Fig.~\ref{fig:figure_001} is utilised to generate the simulated datasets with latent confounders for the experiments in the ``Evaluation with Simulated Data'' subsection of the main text. Code for the g is written in \textit{R}. The details of the data generation process are introduced as follows.	
 
 The simulated datasets are generated from the true DAG in Fig.~\ref{fig:figure_001}, and the specifications of the data generation are as follows: $U, U_1, U_2, U_3, U_4 \sim N(0, 1)$ and $\epsilon_{1, 2, 3, s} \sim N(0, 0.5)$, where $N(,)$ denotes the normal distribution. $X_1 \sim N(0, 1) +0.5* U_2 +\epsilon_{1}$,  $X_2 \sim N(0, 1) +0.5* U_3 +\epsilon_{2}$, $X_3 \sim N(0, 1) +0.5* U_4 +\epsilon_{3}$, $S\sim N(0, 1) + 2*U_1 + 1.5*X_1 + 1.5*X_2 + \epsilon_{s}$,  $X_4 \sim N(1, 1)$,  and $X_5 \sim N(3, 1)$.

 The treatment assignment $T$ is generated from $n$ ($n$ denotes the sample size) Bernoulli trials by using the assignment probability $P(T=1\mid U, U_1, X_3) = [1+exp\{2- 1*U - 1*U_1 -1*X_3\}]$. The potential outcome is generated from $Y_{T} = 2 + 2*T + 2*U + 2*U_3 + 2*U_4 + 1*X_4 + 1*X_5 +\epsilon_{T}$ where $\epsilon_{T}\sim N(0, 1)$. Note that $ACE(T, Y)$ is fixed to 2 on all simulated datasets due to the data generation process.
 
 \subsection{Experiments on three real-world datasets}
 \textbf{Schoolreturning}. The data is from the national longitudinal survey of youth (NLSY), a well-known dataset of US young employees, aged range from 24 to 34~\cite{card1993using}. The treatment is the education of employees, and the outcome is raw wages in 1976 (in cents per hour). The data contains 3,010 individuals and 19 covariates. The covariates include experience (Years of labour market experience), ethnicity (Factor indicating ethnicity), resident information of an individual, age, nearcollege (whether an individual grew up near a 4-year college?), marital status, Father's educational attainment, Mother's educational attainment, and so on. A goal of the studies on this dataset is to investigate the causal effect of education on earnings. Card~\cite{card1993using} used geographical proximity to a college, \ie the covariate \emph{nearcollege} as an instrument variable. We take $\hat{ACE}(T, Y) = 0.1329$\% with 95\% conditional interval (0.0484, 0.2175) from~\cite{verbeek2008guide} as the reference causal effect.

 \textbf{401(k) data}. This dataset is a cross-sectional data from the Wooldridge data sets\footnote{\url{http://www.stata.com/texts/eacsap/}}~\cite{wooldridge2010econometric}. The program participation is about the most popular tax-deferred programs, \ie individual retirement accounts (IRAs) and 401 (k) plans. The data contains 9275 individuals from the survey of income and program participation (SIPP) conducted in  1991~\cite{abadie2003semiparametric}. There are 11 variables about the eligibility for participating in 401 (k) plans, \wrt income and demographic information, including \emph{pira} (a binary variable, \emph{pira} = 1 denotes participation in IRA), \emph{nettfa} (net family financial assets in \$1,000) \emph{p401k} (an indicator of participation in 401(k)), \emph{e401k} (an indicator of eligibility for 401(k)), \emph{inc} (income), \emph{incsq} (income square), \emph{marr} (marital status), \emph{gender}, \emph{age}, \emph{agesq} (age square) and \emph{fsize} (family size). The treatment $T$ is \emph{p401k} and \emph{pira} is the outcome of interest. \emph{e401k} is used as an instrument for $T$ \emph{p401k}~\cite{abadie2003semiparametric}. We take $\hat{ACE}(T, Y) = 0.0712$\% with 95 \% C.I. $(0.047, 0.095)$ from~\cite{abadie2003semiparametric} as the reference causal effect.
 
 \textbf{Sachs}. This data is collected from cell activity measurements for single cell data under a variety of conditions~\cite{sachs2005causal}.  Following the work~\cite{silva2017learning}, we focus on a single condition, \ie simulation with anti-CD3 and anti-CD28. The data contains 853 records and 11 variables~\cite{sachs2005causal}. The treatment is the manipulation of concentration levels of molecule $Erk$. The outcome is the concentration of $Akt$. The other 9 cell products are pretreatment variables~\cite{sachs2005causal}. The data has some weak correlations among variables, but we assume that there are no conditional independencies held between  $Erk$ and the remaining 10 variables. Note that there is not a given instrumental variable. We take the estimated $\hat{ACE}(T, Y) =1.4301$ from the literature~\cite{silva2017learning} (\ie IV.Tetrad's estimated value) as the reference causal effect.

\end{document}